\documentclass[11pt,a4paper]{article}
\usepackage[hyperref]{eacl2021}
\usepackage{times}
\usepackage{latexsym}
\usepackage{graphicx}
\usepackage{hyperref}

\usepackage{microtype}

\aclfinalcopy % Uncomment this line for the final submission
\title{Representations of Language Varieties Are Reliable \\ Given Corpus Similarity Measures}

\author{Jonathan Dunn \\
  Department of Linguistics \\
  University of Canterbury \\
  \texttt{\href{mailto:jonathan.dunn@canterbury.ac.nz}{jonathan.dunn@canterbury.ac.nz}}
}

\date{}

\begin{document}
\maketitle
\begin{abstract}
This paper measures similarity both within and between 84 language varieties across nine languages. These corpora are drawn from digital sources (the web and tweets), allowing us to evaluate whether such geo-referenced corpora are reliable for modelling linguistic variation. The basic idea is that, if each source adequately represents a single underlying language variety, then the similarity between these sources should be stable across all languages and countries. The paper shows that there is a consistent agreement between these sources using frequency-based corpus similarity measures. This provides further evidence that digital geo-referenced corpora consistently represent local language varieties.
\end{abstract}

\section{Introduction}

This paper evaluates whether digital geo-referenced corpora are consistent across registers (web data vs. tweets). This question is a part of validating the use of digital data to represent local language usage in specific places around the world. In other words, recent work has taken samples of varieties like New Zealand English from social media (Twitter) and the web (Common Crawl). But how closely do these two sources actually match up, both with one another and with New Zealand English in offline settings? The first question is about the \textit{reliability} of geo-referenced corpora and the second is about their \textit{validity}. The question of reliability and validity becomes important when we use geo-located digital corpora to model both lexical variation \cite{eosx10, gsg11, kcdsbhskv13, eosx14, donoso-sanchez-2017-dialectometric} and the difference between language varieties \cite{lui-etal-2014-exploring, gpaa16, d18b, dunn-2019-modeling, rangel_rosso_zaghouani_charfi_2020, zampieri_nakov_scherrer_2020}. These computational approaches are quickly refining our understanding of language variation and change. At the same time, though, both of these lines of work depend on large, reliable sources of geographic language data. This paper evaluates these data sources across nine languages and 84 language varieties in order to determine how reliable they are. The data can be explored further alongside demographic information at \href{https://www.earthlings.io/}{\underline{earthLings.io}}.

\section{Reliability and Validity}

If digital geo-referenced corpora are \textit{reliable}, then both web corpora and Twitter corpora from one location should exhibit similar patterns (as quantified by corpus similarity measures). In other words, no matter how many samples we observe of tweets or web pages from New Zealand, we expect a reliable similarity among samples. If digital corpora are not reliable, then it follows that they are not suitable for modelling linguistic variation. This is because a model of a variety like New Zealand English would depend on arbitrary differences in the specific samples being obeserved.

The further question is whether digital geo-referenced corpora are \textit{valid} representations of an underlying population. In other words, if we observed lexical choices by interviewing speakers in New Zealand, we would expect those same lexical choices to be observed in tweets. If these digital corpora are not reliable, then they are also not valid. But reliability itself does not guarantee a valid representation of the underlying population \cite{Nunnally1994}. For example, it is possible that register differences lead to different linguistic behaviour, so that lexical choices in written language may never correspond to lexical choices in spoken language. If this were the case, then it would of course never be possible to generalize from patterns in written language to patterns in spoken language.

\section{Related Work on Validity}

There have been three distinct approaches to the broader problem of validating digital representations of dialects and varieties. First, a \textit{linguistic} approach starts with survey-based dialect studies as a ground-truth. These dialect surveys provide a point of comparison for an underlying variety like American English, one which does not rely on digital language data at all. This line of work has shown that there is a strong correspondence between traditional studies of dialectal variation and both tweets \cite{10.3389/frai.2019.00011} and web-crawled corpora \cite{Cook2017}. This family of results shows that, if we accept dialect surveys as a gold-standard, there is significant justification for substituting digital corpora (like tweets) for much more expensive survey-based data. The correspondence between dialect surveys and tweets, however, is far from perfect. This raises a question: when is the relationship strong enough to justify a purely digital approach to language varieties?

A second \textit{demographic} approach compares digital corpora with ground-truth census data in order to triangulate the similarity between represented populations. For example, given a corpus of tweets from around the world, the geographic density should correspond to some degree with population density \cite{ghg14, Dunn2020}. Otherwise, some locations would be over-represented in the data, thus given more influence. In the same way, a corpus drawn equally from around the world should represent local language use that corresponds to some degree with a census-based study of what languages are used in each country \cite{Dunn2019}. The basic idea here is that geo-referenced corpora which correspond with census data are more likely to represent actual local populations, rather than representing non-local populations like tourists or short-term workers \cite{dunn-etal-2020-measuring}. This line of work has also shown a strong relationship between demographics and digital corpora. But, again, when is that relationship strong enough to justify using geo-referenced corpora alone to model language varieties?

A third \textit{computational} approach leverages some model of language variation to measure relationships between different partitions of linguistic data. For example, recent work has measured the \textit{linguistic similarity} between varieties of English, like New Zealand English vs. Australian English \cite{10.3389/frai.2019.00015,10.3389/frai.2019.00023}. When expanded across aligned corpora, these models can be used to determine if there is a consistent pattern of variation: does New Zealand English have the same distinctive lexical choices on the web that it has in tweets? The basic idea is that if two sources of data (the web and tweets) accurately represent New Zealand English, both should produce similar models of that variety. On the one hand, even if the models agree (\textit{reliability}) it remains possible that both sources of data are distorting the underlying population (\textit{validity}). But it is more likely that such reliability confirms the validity of a purely digital approach to language varieties. This kind of work has also been systematically evaluated across different similarity measures, with a ground-truth derived from studies in perceptual dialectology \cite{Heeringa2002, Heeringa2006}.

This paper expands on these three approaches by systematically experimenting with the reliability of geo-referenced data (i) across register boundaries, (ii) across geographic boundaries, and (iii) across language boundaries. The more reliable the relationship between web data and tweets, the more confidence we can have in computational models of language varieties and linguistic variation.

\section{Measuring Corpus Similarity}

The approach in this paper is to measure where digital geo-located corpora diverge from one another, across languages and across countries. Unlike previous work, this does not rely on a comparison with an external ground-truth (i.e., dialect surveys, census data, or pre-trained models of variation). First, we collect geo-referenced web data and tweets across 84 varieties of nine languages. Each variety is represented using aligned corpora drawn from web pages and from tweets. These representations are distributed across many individual samples of 1 million words each. 

This design allows us to make a direct comparison between samples that have been drawn from the same varieties. Thus, for example, we can measure the similarity of a single variety across register boundaries: New Zealand English on the web and in tweets. And we can also measure the similarity between varieties within a single register: New Zealand English and Australian English in tweets. This is related, for example, to the problem of measuring similarity within a single language variety over time \cite{pichel-campos-etal-2018-measuring}. The ultimate goal of the experiments described in this paper is to examine the reliability of digital representations of language varieties.

\begin{table}
\centering
\begin{tabular}{|c|c|c|c|}
\hline
\textbf{Lang.} & \textbf{Varieties} & \textbf{Samples \textsc{tw}} & \textbf{Samples \textsc{cc}} \\
\hline
ara & 13 & 185 & 150 \\
deu & 7 & 76 & 124 \\
eng & 14 & 259 & 262 \\
fra & 15 & 186 & 283 \\
ind & 3 & 45 & 56 \\
nld & 2 & 43 & 46 \\
por & 4 & 53 & 71 \\
rus & 9 & 99 & 216 \\
spa & 17 & 336 & 322 \\
\hline
\textbf{Total} & \textbf{84} & \textbf{1,282} & \textbf{1,530} \\
\hline
  \end{tabular}
  \caption{Varieties and Samples by Language}
  \label{tab:1}
\end{table}

We start by drawing on measures of corpus similarity, which calculate the distance between two corpora \cite{Kilgarriff2001,kilgarriff-rose-1998-measures}. Previous work has shown that frequency-based measures out-perform more sophisticated approaches that make use of language models or topic models \cite{fothergill-etal-2016-evaluating}. The highest-performing frequency models work by ranking word or character frequencies and using either the Spearman $rho$ or the $\chi_2$ measure between frequency ranks. The $\chi_2$ performs slightly better for samples with the same number of features, but the Spearman $rho$ has more flexibility for dealing with unattested features while experiencing only a minimal impact on performance \cite{Kilgarriff2001}.

Based on these previous evaluations, we rely on two variants of the Spearman frequency-based similarity measure: one variant uses unigram frequencies and one variant uses character trigram frequencies (motivated by related work on language identification). The experiments in this paper are performed using equal-sized samples, where each sample contains 1 million words. These two measures are labelled as \textit{Word} and \textit{Char} in the tables that follow. The inventory of features (i.e., words or character trigrams) consists of the most common 100k features across all samples from a given language. A higher Spearman value indicates a higher similarity between samples.

In the ideal case, there would be a consistent level of similarity across samples representing all languages and all countries. In other words, we would see a similar impact of register variation (web vs. tweets) regardless of which specific language variety is involved. A result like this would suggest that both sources of geo-located language are representing a single underlying linguistic object (i.e., offline New Zealand English). 

The alternate outcome is that there is an arbitrary fluctuation between web usage and tweet usage. For example, it could be the case that New Zealand English is very similar on the web and in tweets, but Australian English shows a wide divergence between the two, with different lexical choices being represented. If this were the case, the digital representation of New Zealand English would be more reliable than the representation of Australian English. This work is related to previous explorations of \textit{noise} in social media \cite{baldwin-etal-2013-noisy}, but with a focus on the interaction between geographic variation and register variation. The question is: how consistent are language varieties when sampled from different digital sources?

\section{Data and Experimental Design}

The experiments draw on existing geographic corpora from the web (ultimately derived from the Common Crawl data) and from tweets \cite{dunn-adams-2020-geographically,Dunn2020}. Both of these data sources are aligned by language and country: for example, each variety like Cuban Spanish or New Zealand English is present from both sources of data. The experiments include nine languages: Arabic (ara), German (deu), English (eng), French (fra), Indonesian (ind), Dutch (nld), Portuguese (por), Russian (rus), and Spanish (spa). These languages each have a different number of varieties, as shown in Table 1, ranging from just 2 (nld) to 17 (spa). In this table, \textit{Varieties} refers to the number of national dialects or varieties (like New Zealand English) that are represented. \textit{Samples} refers to the total number of observations for that language, across all varieties. Finally, \textit{TW} refers to tweets and \textit{CC} to web data.

Each sample contains 1 million words. Some varieties are very well represented, with hundreds of samples (and thus hundreds of millions of words); but other varieties are less well represented, with only a few samples. We restrict the maximum number of samples per variety per register to 20. As shown in Table 2 for Spanish, there is variation in the number of samples per data source. For example, some countries like Mexico have many samples for both tweets and web data (20 and 20, respectively, because of the cap). But other countries are imbalanced: for example, the Dominican Republic has 20 samples from the web but only 4 drawn from tweets. Overall, these experiments are based on approximately 1.2 billion words from tweets and 1.5 billion words from web corpora.

\begin{table}
\centering
\begin{tabular}{|l|c|c|c|}
\hline
\textbf{Country} & \textbf{Samples \textsc{tw}} & \textbf{Samples \textsc{cc}} \\
\hline
Argentina & 20 & 16 \\
Bolivia & 20 & 16 \\
Chile & 20 & 20 \\
Colombia & 20 & 20 \\
Costa Rica & 20 & 20 \\
Cuba & 20 & 20 \\
Dom. Rep. & 4 & 20 \\
Ecuador & 20 & 20 \\
El Salvador & 20 & 20 \\
Guatemala & 20 & 20 \\
Honduras & 20 & 20 \\
Mexico & 20 & 20 \\
Paraguay & 20 & 2 \\
Peru & 20 & 20 \\
Spain & 20 & 20 \\
United States & 12 & 20 \\
Uruguay & 20 & 19 \\
Venezuela & 20 & 9 \\
\hline
  \end{tabular}
  \caption{Number of Samples by Variety for Spanish}
  \label{tab:2}
\end{table}

The question behind the experiments in this paper is whether there is a consistent relationship between corpora drawn from the web and from tweets, across many different language varieties. If there is a significantly stable relationship, this increases our confidence that digital corpora accurately represent the underlying geographic language varieties. In other words, the stable relationship would be a result of two different sources that are measuring the same real-world entity (e.g., a single language variety). But, in contexts where this relationship breaks down, we can have less confidence in the validity of digital corpora for modelling language varieties. We present five sets of experiments, each focused on a specific part of this problem. 

The first experiment (Section 6), focuses on the internal consistency of each variety-register combination: given similarity relationships between many samples of Mexican Spanish tweets, how stable is the internal similarity between samples? The basic idea is that any variety-register combination with a low internal similarity is less reliable. 

The second experiment (Section 7) evaluates the accuracy of both word-frequency and character-frequency measures. The idea in these experiments is to validate the underlying similarity measure to ensure that the later experiments are also valid. 

The third experiment (Section 8) is the main part of the paper: measuring the difference between registers (web vs. tweets) for each language variety. Our hypothesis is that only a significantly stable relationship supports the use of digital corpora for representing language varieties. 

The fourth experiment (Section 9) looks for differing results by language. It may be the case, for instance, that varieties of English and Spanish are adequately represented by digital corpora, but not varieties of Dutch.

The fifth experiment (Section 10) creates an \textit{average} frequency vector for each language across all varieties and registers. This average is used to situate each national variety according to its relative distance from that center, creating a rank of which varieties are the most divergent or the most unique. There is one rank for each register and the experiment tests where there is a significant relationship between ranks for each language. Taken together, this set of experiments is designed to evaluate the suitability of digital geo-referenced corpora for representing language varieties, building on the validation work discussed in Section 3.

\section{Experiment 1: Consistency Across Samples}

The first question is the degree to which each language variety-register combination (e.g., Mexican Spanish tweets) is internally consistent. We take 50 observations of unique pairs of samples, each representing the same variety-register combination. Then we measure the similarity between the two samples and look at the distribution of similarity values for each variety-register combination. Varieties with a high average similarity are more internally consistent than varieties with a low average similarity.

The results are shown in Table 3 by language, divided into \textit{Word} and \textit{Character} features as discussed in Section 4. Data from tweets (\textit{TW}) is shown on the left and from the web (\textit{CC} for Common Crawl) is shown on the right. These measures show the stability by language, where each value is drawn from approximately 50 random comparisons between different samples representing the same context. As throughout this paper, higher values indicate higher similarity (here, meaning more internal similarity).

First, this table shows that \textit{character} representations are more consistent across languages, although the accuracy of each type of feature is not evaluated until Section 7. In other words, the average internal similarity for tweets ranges from 0.82 to 0.85 for character features but from 0.63 to 0.71 for word features. At the same time, a smaller range of variation also could mean that character features are less able to discriminate between varieties. 

\begin{table}
\centering
\begin{tabular}{|c|c|c||c|c|}
\hline
\textbf{Language} & \textbf{TW} & \textbf{TW} & \textbf{CC} & \textbf{CC} \\
~ & \textbf{Word} & \textbf{Char} & \textbf{Word} & \textbf{Char} \\
\hline
ara & 0.71 & 0.84 & 0.70 & 0.76 \\
deu & 0.66 & 0.85 & 0.57 & 0.72 \\
eng & 0.68 & 0.83 & 0.63 & 0.71 \\
fra & 0.68 & 0.82 & 0.64 & 0.72 \\
ind & 0.71 & 0.83 & 0.64 & 0.73 \\
nld & 0.63 & 0.82 & 0.53 & 0.71 \\
por & 0.67 & 0.82 & 0.64 & 0.72 \\
rus & 0.70 & 0.83 & 0.60 & 0.71 \\
spa & 0.65 & 0.82 & 0.61 & 0.71 \\
\hline
  \end{tabular}
  \caption{Consistency Within Language Varieties}
  \label{tab:3}
\end{table}

Second, this table shows that tweets from a particular variety have a higher internal similarity than web data from that variety. This is true for all languages and for both word and character features. While this experiment does not indicate \textit{why} this is the case, we can offer some potential reasons: (i) it could be that tweets are more similar in terms of discourse functions, while web data represents a range of different communication purposes; (ii) it could be that tweets represent a more restricted subset of local populations (e.g., Twitter users), which have been shown to be more influenced by demographic attributes like per capita GDP. 

Third, all of these Spearman correlations are highly significant, which means that there is at least a strong relationship between representations of each variety drawn from the web and from tweets. At the same time, as discussed in Section 3 in regards to alternate methods of evaluation, it is not clear what threshold for similarity is sufficient. This question is what the remainder of the experiments attempt to address.

\section{Experiment 2: Word vs Character Features}

The next question is two-fold: First, which frequency-based measure of similarity is more reliable in this setting: words or characters? Previous work on corpus comparison has largely conducted evaluations using only English data and has not included multiple language varieties. Second, regardless of which feature is best, how reliable are the measures? Because the samples are each 1 million words, this is a significantly easier problem than identifying varieties in small samples. 

In this experiment we look at the average accuracy for each language for each type of feature. For each variety, we take 100 pairs that come from the same register (tweets against tweets) and 100 pairs that come from different registers (tweets against web pages). If the similarity for a pair is closest to the average similarity for same-register pairs, we predict that each sample comes from the same register (and so on for cross-register pairs). This is a simple framework for evaluating each measure: the percent of correct predictions. This is similar to previous evaluation methods for corpus similarity measures \cite{Kilgarriff2001}.

\begin{table}
\centering
\begin{tabular}{|c|c|c|c|}
\hline
\textbf{Language} & \textbf{Word} & \textbf{Character} \\
\hline
ara & 99.0\% & \textbf{99.1\%} \\
deu & 95.7\% & \textbf{98.9\%} \\
eng & 93.9\% & \textbf{98.9\%} \\
fra & 98.1\% & \textbf{99.8\%} \\
ind & 100.0\% & 100.0\% \\
nld & 94.3\% & \textbf{100.0\%} \\
por & \textbf{98.4\%} & 96.3\% \\
rus & 94.7\% & \textbf{98.7\%} \\
spa & 96.7\% & \textbf{97.5\%} \\
\hline
  \end{tabular}
  \caption{Feature Evaluation by Accuracy}
  \label{tab:4}
\end{table}

The results, shown in Table 4, have a high accuracy in general, as we expect given that each sample is 1 million words. The important question is which similarity measure has the highest accuracy: in every case except for Portuguese the character features are more accurate. For simplicity, we report only the character-based measure in the remainder of the experiments. This experiment also provides a ground-truth confirmation that the measures are meaningful on this data.

\begin{figure}[htp]
    \centering
    \includegraphics{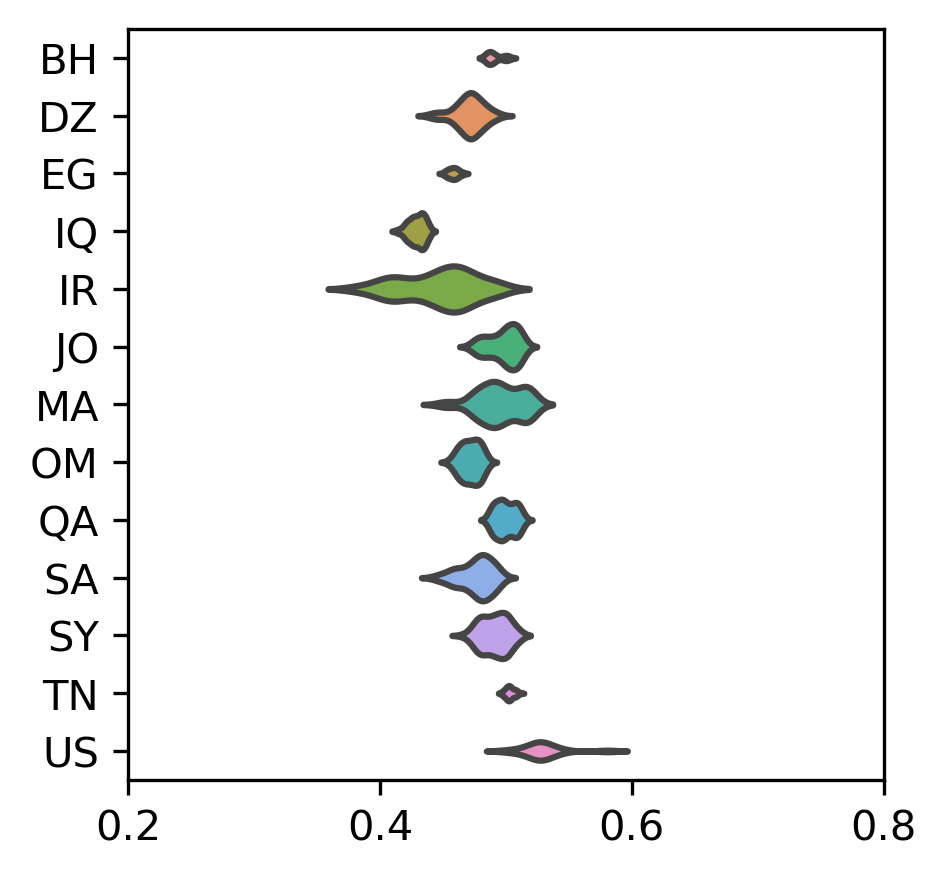}
    \caption{Cross-Register Similarity (Arabic)}
    \label{fig:ara}
\end{figure}

\begin{figure}[htp]
    \centering
    \includegraphics{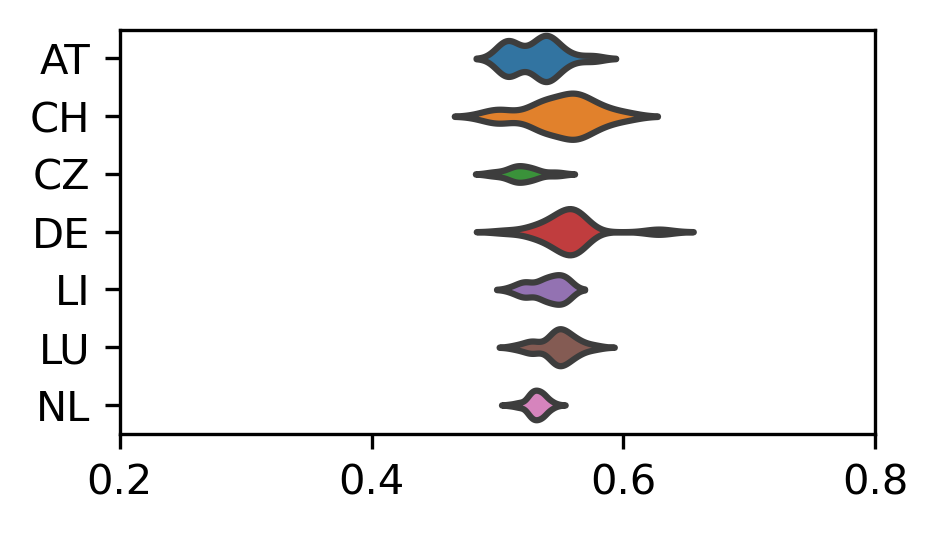}
    \caption{Cross-Register Similarity (German)}
    \label{fig:deu}
\end{figure}

\begin{figure}[htp]
    \centering
    \includegraphics{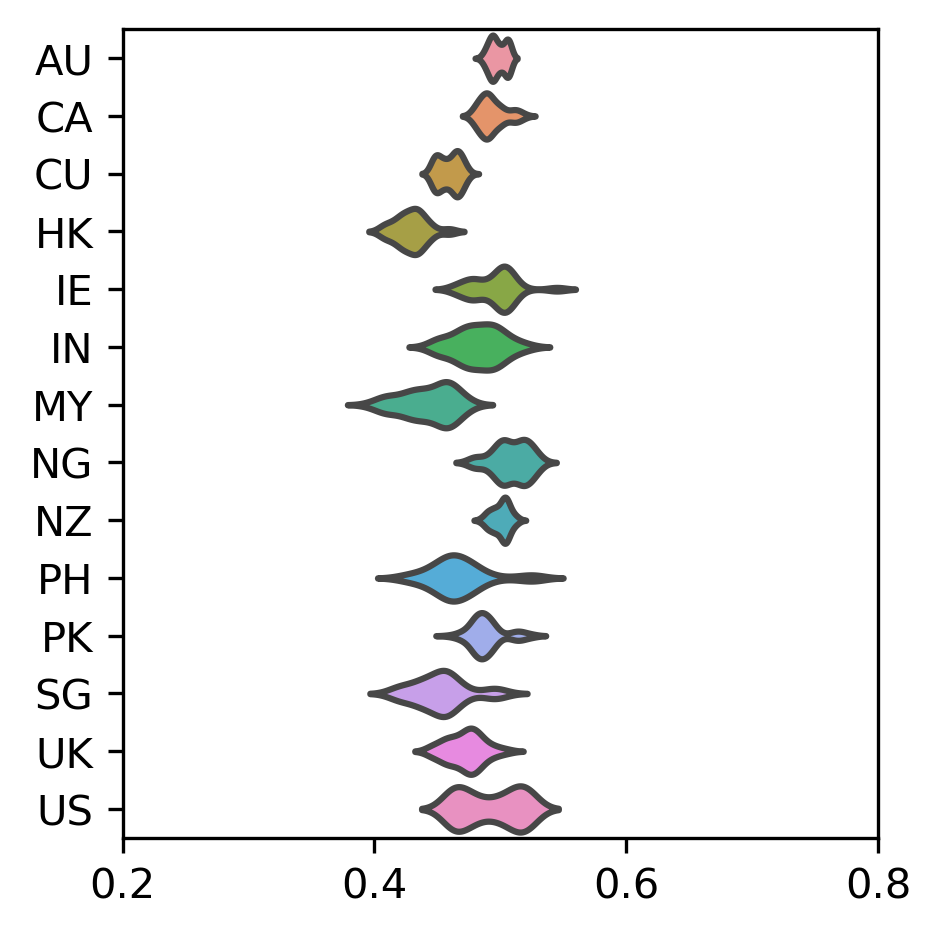}
    \caption{Cross-Register Similarity (English)}
    \label{fig:eng}
\end{figure}
\section{Experiment 3: Differences in Register}

Now that we have established that there is a significant relationship between registers across all varieties and that the character-based measure is generally more accurate, we proceed to the main question: how robust is the relationship between registers? There are two competing hypotheses here: First, it could be the case that one or both sources of digital data only poorly represent a given language variety. In this case, there would be wide fluctuations in similarity values across varieties and across languages (because the similarity would not be driven by shared populations and shared linguistic patterns). Second, it could be the case that both sources of digital data consistently represent a given language variety. Because of register-based variation, the similarity between sources may not be perfect; but it should be consistent. 

We use the samples described above to create a population of cross-register pairs, where each pair consists of two samples from the same variety. For instance, we take 100 random comparisons of Mexican Spanish with other Mexican Spanish samples from a different source (i.e., web vs tweets). This provides a population of between-register scores for each language variety. If the distance between registers is consistent across all contexts, this is evidence for the second hypothesis: that both sources of data are related representations of a single underlying language variety.

The first set of results, for Arabic, is shown in Figure 1 as a violin plot of the distribution of cross-register similarity scores for each country. In the case of Arabic, the average similarity by country ranges from 0.43 (Iraq, IQ) to 0.53 (United States, US). Note the combination here of native varieties and non-native or immigrant varieties. This violin plot shows the distribution of scores, most of which are contained within a tight central band (i.e., are relatively consistent across varieties). Arabic in Iran (IR) and Morocco (MA) have broader distributions, showing that there is more variety within samples from these countries. The US, a non-native variety, has outliers, resulting in thin lines on either side of the plot. There are also differing numbers of samples by country, so that Egypt (EG) and Tunisia (TN) are located in the center but, with few samples, have a small footprint.

Moving on to German, in Figure 2, we see a similar pattern: the distribution (with somewhat higher scores) is centered within the same values across countries. What matters for the hypothesis in this paper is the range of values for cross-register similarity, here with averages ranging from 0.52 (Czechia, CZ) to 0.56 (Germany, DE). This constrained range indicates that there is a stable relationship between digital sources for these varieties.

\begin{figure}[htp]
    \centering
    \includegraphics{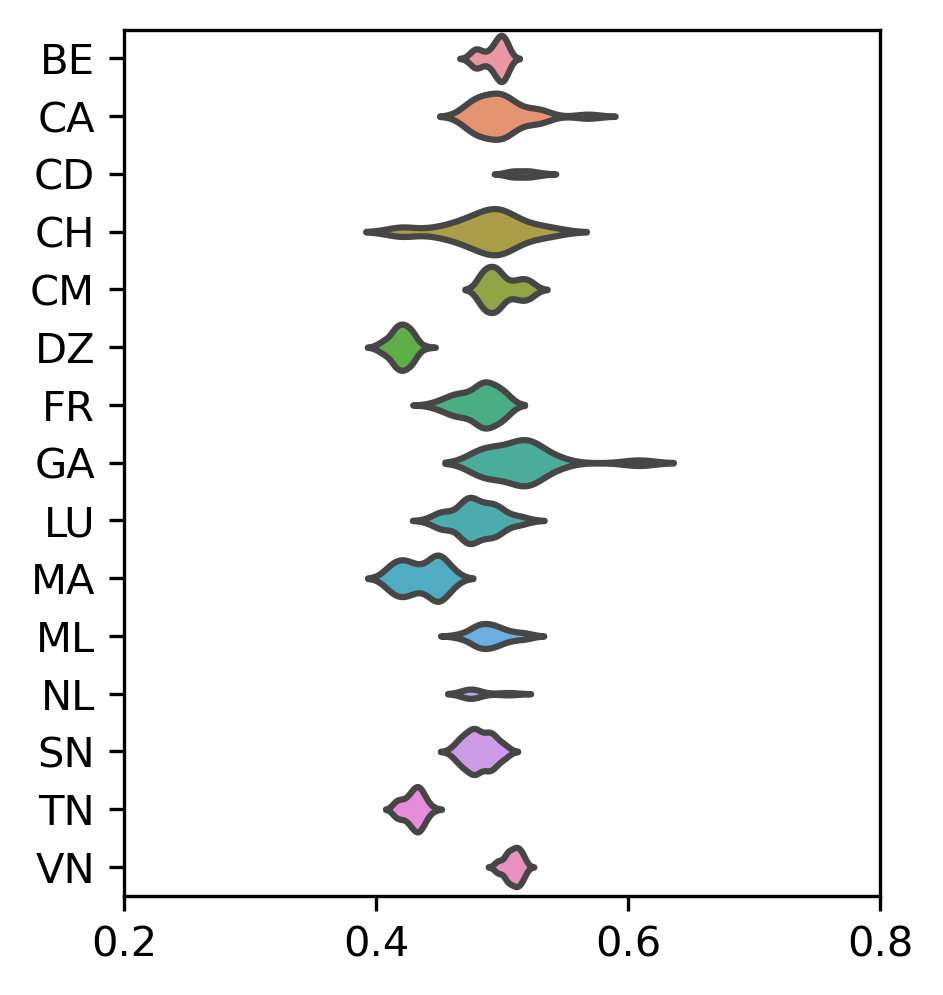}
    \caption{Cross-Register Similarity (French)}
    \label{fig:fra}
\end{figure}

We next evaluate varieties of English (Figure 3) and French (Figure 4). Both of these languages are more widely used in digital contexts, thus having a larger number of varieties. English ranges from an average of 0.43 (Hong Kong, HK) to 0.51 (Nigeria, NG). French ranges from an average of 0.42 (Algeria, or DZ) to 0.52 (Gabon, GA, and the Democratic Republic of Congo, CD). In both languages we see a somewhat wider range, with outliers like Hong Kong English or Algerian and Tunisian French; but still the gap between registers remains largely consistent for most varieties.

\begin{figure}[htp]
    \centering
    \includegraphics{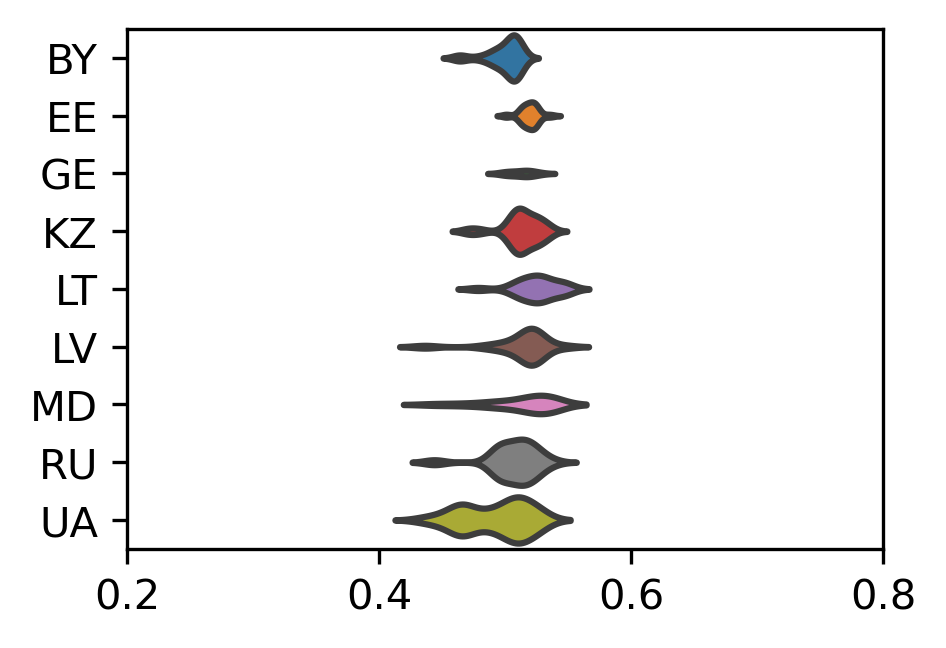}
    \caption{Cross-Register Similarity (Russian)}
    \label{fig:rus}
\end{figure}

The results for Russian (Figure 5) and Spanish (Figure 6) continue the same pattern of relative consistency. The figures for Dutch, Indonesian, and Portuguese are excluded because they each have a small number of varieties that largely overlap. The figures for Russian and Spanish show greater consistency across countries, although Russian does have a few outliers for each country (i.e., the narrow tails). Russian averages range from 0.49 (Ukraine, UA) to 0.52 (Estonia, EE). Spanish shows a slightly larger range of averages, from 0.48 (Argentina, AR) to 0.55 (El Salvador, ES). Once again, however, the relationship between registers remains stable.

\begin{figure}[htp]
    \centering
    \includegraphics{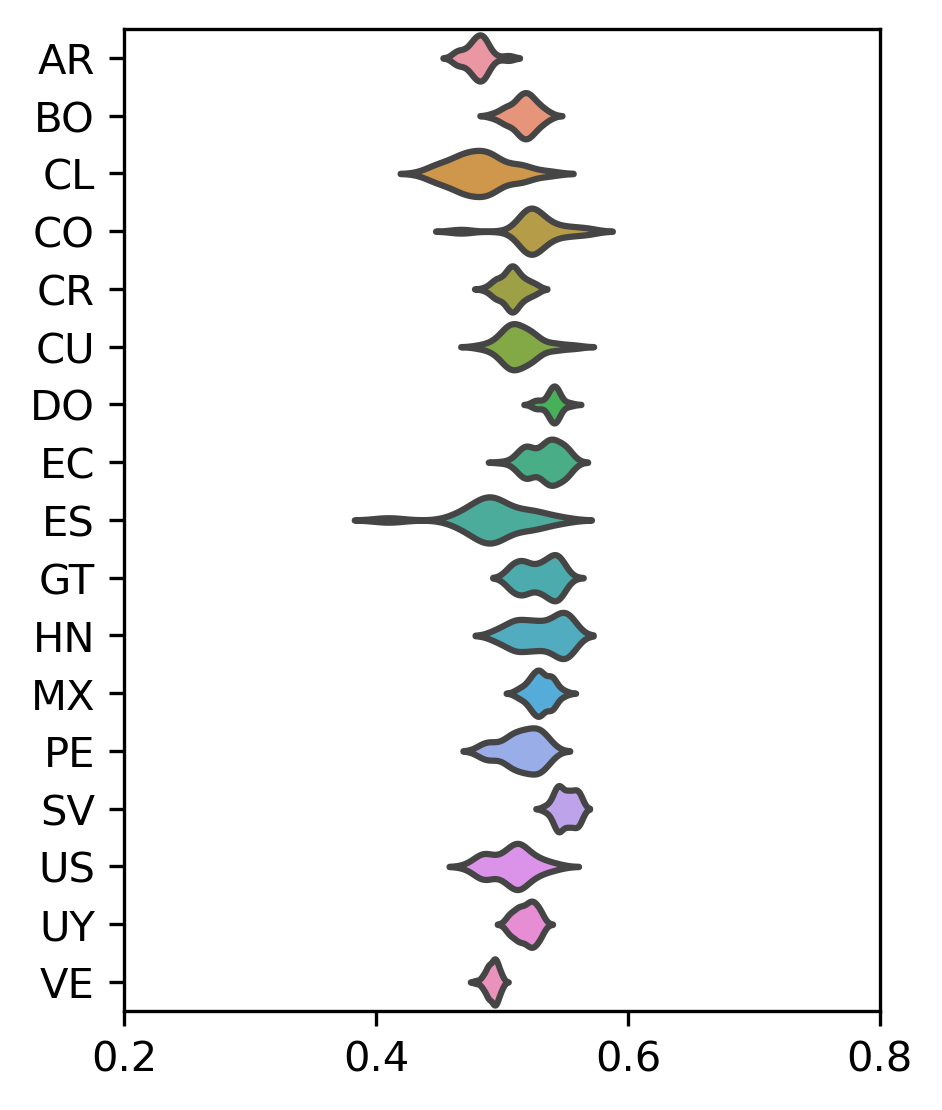}
    \caption{Cross-Register Similarity (Spanish)}
    \label{fig:spa}
\end{figure}

\begin{table*}[t]
\centering
\begin{tabular}{|c|c|c|c|c|}
\hline
\textbf{Language} & \textbf{TW} & \textbf{TW} & \textbf{CC} & \textbf{CC}\\
~ & Same Varieties & Different Varieties & Same Varieties & Different Varieties \\
\hline
ara & 0.84 & 0.81 & 0.76 & 0.67 \\
deu & 0.85 & 0.83 & 0.72 & 0.68 \\
eng & 0.83 & 0.80 & 0.71 & 0.66 \\
fra & 0.82 & 0.77 & 0.72 & 0.66 \\
ind & 0.83 & 0.80 & 0.73 & 0.68 \\
nld & 0.82 & 0.80 & 0.71 & 0.69 \\
por & 0.82 & 0.80 & 0.72 & 0.64 \\
rus & 0.83 & 0.80 & 0.71 & 0.68 \\
spa & 0.82 & 0.80 & 0.71 & 0.67 \\
\hline
  \end{tabular}
  \caption{Variation Within and Across Countries}
  \label{tab:12}
\end{table*}

The results in this section suggest that both web data and tweets provide reliable representations for a large number of language varieties, regardless of the specific language or country involved. The distance between samples from each source is stable across both categories (language and country). This pattern is what we expect if both sources adequately represent the underlying language variety. This reliability converges with the previous work discussed in Section 3 to suggest that digital geo-referenced corpora do in fact represent local language use. As with previous evaluations, however, it is difficult to define a threshold for reliability.

\section{Experiment 4: Differences by Language}

Our next set of experiments further explores the interaction between dialect variation (e.g., New Zealand vs. Australian English) and register variation (e.g., web data vs. tweets). As we compare a large number of samples for each language, how much of the variation comes from register-effects and how much from geographic-effects? The experiments shown in Table 5 compare the similarity of samples drawn from the same variety (thus controlling for geographic effects) with the similarity of samples drawn from different varieties (thus adding geographic effects). The additional similarity of within-variety comparisons can be attributed to geographic variation. 

For example, Arabic tweets have 0.03 increased similarity when compared within varieties instead of between varieties. The difference for web data is larger in most cases. We test for significance using a t-test with 50 observed pairs in each condition for each language. In each case there is a highly significant difference. This confirms that we are observing geographic variation across varieties in addition to the register variation that is quantified in the previous section. 

This geographic variation is a secondary finding, however, in the sense that much more precise methods of modelling geographic variation exist. The purpose here is simply to confirm that each language shows a significant geographic effect given this measure applied to these sources of data.

\section{Experiment 5: Relations Between Varieties}

Our final experiment looks at relationships between national varieties or dialects. A final piece of evidence comes from ranks of varieties: if both the web and tweets make Singapore English most similar to Malaysian English, this confirms the reliability digital sources of geo-referenced data. 

First, we create an average representation across all samples for a given language using the mean feature frequency. Second, we estimate the average feature frequency for each variety-register combination (like Mexican Spanish tweets). Third, we compare the similarity between (i) the language-based mean feature value and (ii) the average for a specific variety (like Mexican Spanish tweets). 

This provides a rank of distance from the average frequency for each variety. For example, what is the most unusual variety of English? We do this for both registers, \textit{TW} and \textit{CC} in Table 6. The question is, do these two sources of data agree about relationships \textit{between} dialects as well as between registers? The Spearman rank correlations are used to measure how well each register ranks varieties in the same manner.

\begin{figure*}[t]
  \includegraphics[width=460pt]{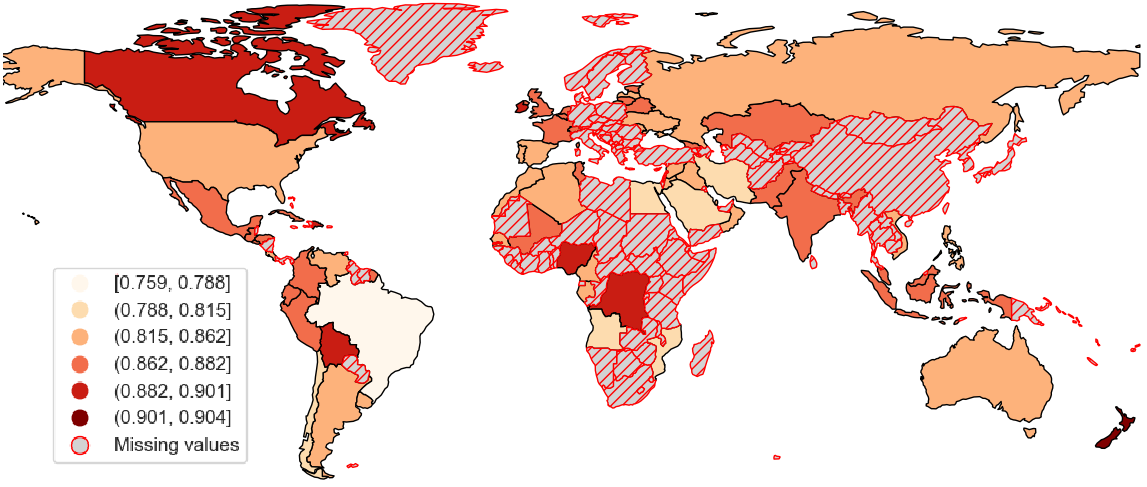}
  \caption{Average cross-domain similarity by country.}
  \label{fig:map1}
\end{figure*}

The results for all languages taken together is highly significant, with a correlation of 0.453. Taken individually, some languages are also significant: German, French, Portuguese, and Russian. Given the small number of observations when constrained to individual languages, the lack of significance at that level is not particularly important. What this shows, however, is that even a simple similarity measure can produce, overall, significantly related ranks of varieties. The importance is not the specific ranks, because we know that better methods for modelling language varieties exist. The importance is that we have additional evidence for stability in the relationship between registers and varieties. This is precisely what we would see if both sources of digital data were valid representations of a single underlying linguistic object.

\begin{table}
\centering
\begin{tabular}{|c|c|c|}
\hline
\textbf{Language} & \textbf{Spearman $rho$} & \textbf{Significance} \\
\hline
\textbf{All} & \textbf{0.453} & \textbf{*} \\
\hline
ara & 0.291 & NS \\
deu & 0.785 & \textbf{*} \\
eng & 0.428 & NS \\
fra & 0.561 & \textbf{*} \\
ind & 0.399 & NS \\
nld & 0.799 & NS \\
por & 0.899 & \textbf{*} \\
rus & 0.797 & \textbf{*} \\
spa & 0.051 & NS \\
\hline
  \end{tabular}
  \caption{Correlation between rankings by register}
  \label{tab:13}
\end{table}

\section{Geographic Trends}

The focus so far has been on varieties of languages. A different way of framing the question of register similarity is by country: is there a geographic pattern to where the web and tweets are more or less similar? Here we take cross-register similarity values as described above in Section 8. When we bring all languages together, these experiments represent 66 countries. Of these, 10 are represented by two or more languages; the map in Figure 7 averages the values by country to show geographic trends in cross-register similarity.

When we focus on countries, the average cross-register similarity value is 0.855. Most countries fall with the range of 0.80 to 0.89 (62 out of 66). Only two countries fall below 0.80, meaning that they show a low agreement between registers: Bahrain and Brazil. And only two countries are above 0.90, meaning that they show a higher agreement between registers: Nigeria and New Zealand. Note that the lower value for Brazil likely results from the lower accuracy of character-based measures for Portuguese (c.f., Table 4).

This kind of geographic analysis does not show larger trends. For example, one common trend in digital corpora is that wealthy western countries like the US, Canada, Western Europe, and Australia and New Zealand pattern together. But this is not the case here, with Canada and New Zealand contrasting with the US and Australia. Thus, a geographic analysis further confirms the reliability of geo-referenced corpora in the sense that what little variation we do see does not correspond with expected geographic patterns.

\section{Conclusions}

The experiments in this paper have shown that there is a consistent relationship between the representation of language varieties across two sources of geo-referenced digital corpora: the web and tweets. In other words, these digital corpora are \textit{reliable}. This finding is important because it disproves the alternate hypothesis that one or both sources is a poor or arbitrary representation of the underlying language varieties. This kind of unreliability would lead to an inconsistent relationship between registers. Reliability is not the same as validity. However, the reliability shown here converges with evidence from dialect surveys and census data and computational models of language varieties, all of which do evaluate the validity of digital corpora. 

This robust reliability gives us confidence in the use of digital corpora to study linguistic variation. And this evaluation of both reliability and validity is important because it justifies the use of computational modelling which, in turn, enables global-scale experiments that help us better understand language variation and change.

\bibliography{eacl2021}
\bibliographystyle{acl_natbib}

\end{document}